\documentclass[conference]{IEEEtran}
\IEEEoverridecommandlockouts
\usepackage{cite}
\usepackage{amsmath,amssymb,amsfonts}
\usepackage{algorithmic}
\usepackage{graphicx}
\usepackage{textcomp}
\usepackage{xcolor}
\usepackage{mathtools}
\usepackage[bottom]{footmisc}
\usepackage[square,numbers,sort&compress]{natbib}
\usepackage{breqn}
\usepackage{color}
\usepackage{threeparttable}
\usepackage[normalem]{ulem}
\usepackage{comment}
\newcommand{\minimize}[2]{\ensuremath{\underset{\substack{{#1}}}{\textrm{minimize}}\;\;#2 }}

\def\BibTeX{{\rm B\kern-.05em{\sc i\kern-.025em b}\kern-.08em
    T\kern-.1667em\lower.7ex\hbox{E}\kern-.125emX}}
\begin{document}

\title{ConFuse: Convolutional Transform Learning Fusion Framework For Multi-Channel Data Analysis\\
\thanks{This work was supported by the CNRS-CEFIPRA project
under grant NextGenBP PRC2017.}
}

\author{\IEEEauthorblockN{1\textsuperscript{st} - Pooja Gupta}
\IEEEauthorblockA{\textit{Computer Science and Engineering} \\
\textit{Indraprastha Institute of Information Technology Delhi}\\
Delhi, India \\
poojag@iiitd.ac.in}
\and
\IEEEauthorblockN{2\textsuperscript{nd} - Jyoti Maggu}
\IEEEauthorblockA{\textit{Computer Science and Engineering} \\
\textit{Indraprastha Institute of Information Technology Delhi}\\
Delhi, India \\
jyotim@iiitd.ac.in
}
\and
\IEEEauthorblockN{3\textsuperscript{rd} - Angshul Majumdar}
\IEEEauthorblockA{\textit{Electronics and Communications Engineering} \\
\textit{Indraprastha Institute of Information Technology Delhi}\\
Delhi, India \\
angshul@iiitd.ac.in}
\and
\IEEEauthorblockN{4\textsuperscript{th} - Emilie Chouzenoux}
\IEEEauthorblockA{\textit{CVN, Inria Saclay} \\
\textit{Univ. Paris-Saclay}\\
Gif-sur-Yvette, France \\
emilie.chouzenoux@centralesupelec.fr}
\and
\IEEEauthorblockN{5\textsuperscript{th} - Giovanni Chierchia}
\IEEEauthorblockA{\textit{LIGM, CNRS, ESIEE Paris} \\
\textit{Univ. Gustave Eiffel}\\
Marne-la-Vallée, France \\
giovanni.chierchia@esiee.fr}
}

\maketitle

\begin{abstract}
This work addresses the problem of analyzing multi-channel time series data %. In this paper, we 
by proposing an unsupervised fusion framework based on %the recently proposed 
convolutional transform learning. Each channel is processed by a separate 1D convolutional transform; the output of all the channels are fused by a fully connected layer of transform learning. The training procedure takes advantage of the proximal interpretation of activation functions. We apply the developed framework to multi-channel financial data for stock forecasting and trading. We compare our proposed formulation with benchmark deep time series analysis networks. The results show that our method yields considerably better results than those compared against. 
\end{abstract}


\section{Introduction}
\label{sec:intro}
Several real-world problems can be expressed as a multi-channel time series analysis. Consider for instance the problem of demand forecasting \cite{ref2}, which consists of estimating the power consumption at a future point given the available information until the current instant. In building-level forecasting \cite{ref3}, typical inputs are power consumption, temperature, humidity, and occupancy, which can be considered as separate channels of a time series. Biomedical signal processing also involves multi-channel sequences, for instance in the problem of blood pressure (systolic and diastolic) estimation, the inputs are usually both the electrocardiogram (ECG) and the pulse pleithismogram (PPG) signals, forming three channels (2 for ECG and 1 for PPG) \cite{ref4}. The last example, that we will particularly look at in this work, is stock price forecasting, where the next day close price must be predicted from five inputs of the current day. 

Until a few years back, the standard deep learning approach for modeling time series data was based on long short-term memory (LSTM) \cite{ref5} or gated recurrent unit (GRU) \cite{ref6}. Although theoretically appropriate, LSTM and GRU are hardware intensive and more time and memory consuming than 1D CNN. Moreover, LSTM and GRU could not model very long sequences. Owing to these shortcomings, 1D CNNs are gradually replacing LSTM, GRU and other RNN variants\footnote{https://towardsdatascience.com/the-fall-of-rnn-lstm-2d1594c74ce0}. Another class of methods convert time series data to a matrix form and use 2D CNNs instead \cite{ref10, ref11, ref12}. This 2D CNN model is especially popular in financial data analysis \cite{ref1,ref13,ref14}.

Deep learning has also been widely used for analyzing multi-channel / multi-sensor signals. In several such studies, all the sensors are stacked one after the other to form a matrix and 2D CNN is used for analyzing these signals \cite{ref7,ref8}. It is worth mentioning that most existing works address inference from multi-channel data using a supervised end-to-end machine learning paradigm \cite{ref7,ref8,ref9,ref10,ref11,ref12,ref13,ref14}, where the input is the raw multi-channel signal and the output is the inferred parameter (e.g., class label, regression value). A recent alternative has been proposed in \cite{Lowe_NIPS2019}, relying on an unsupervised method that learns feature representations from multi-channel data, and then uses these features as an input of a suitable classifier or regression method, adapted to the user's need. Such an unsupervised strategy may offer greater flexibility than supervised ones. Consider for instance the problem of ECG data classification, involving either four or sixteen classes~\cite{NatureMed2019}. In the said work, two distinct deep neural networks are trained to solve both cases. In contrast, if the learning paradigm was unsupervised, only one network would be needed, providing as an output relevant features, which could, later on, serve as input to a third-party classifier or regressor. 

In this work, we focus on two important problems of stock analysis, namely stock forecasting and stock trading. The former is a regression problem where the task is to predict the value of a stock, and the latter amounts to decide whether to buy or sell a stock. Instead of creating two separate end-to-end models, for performing classification and regression, we propose to learn a single unsupervised model. Motivated by the success of 1D CNNs for time series processing, coupled with the need for an unsupervised representation learning tool, we propose to make use of our recently introduced convolutional transform learning (CTL) approach \cite{ref1}. Each channel is hence processed by 1D CTL and the resulting representations are fused by a fully connected network of transform learning, leading to the so-called \emph{ConFuse} approach. An original end-to-end training strategy is proposed, that takes advantage of a key property of activation functions, thus allowing the use of the efficient stochastic procedure for learning the architecture parameters. Comparisons with state-of-the-art methods illustrate the benefits of the proposed method.

The rest of the paper will be structured as follows. Section 2 introduces the CTL representation learning machinery and presents the proposed method \emph{ConFuse} for unsupervised multi-channel fusion. Experimental results are provided in Section 3. Conclusions are discussed in Section 4.

\section{PROPOSED FORMULATION}
\label{sec:proposed}

As already mentioned in the introduction, our approach consists in introducing a novel unsupervised framework for multi-channel data representation learning. A key ingredient of the latter will be the recently introduced CTL~ \cite{ref1}. For the sake of clarity, we recall first the main steps of the CTL technique. Then, we will present the overall \emph{ConFuse} architecture and discuss its training.

\subsection{Convolutional transform learning}
CTL aims at learning a set of $M$ convolutional filters $(t_m)_{1 \leq m \leq M}$ from $K$ observed samples $(s^{(k)})_{1 \leq k \leq K}$, so as to generate a set of $MK$ feature vectors $(x_m^{(k)})_{1 \leq k \leq K, 1 \leq m \leq M}$ representing adequately the data. The representation learning model is expressed as
\begin{equation*}
\forall m \in \{ 1, \ldots,M\},\; \forall k \in \{ 1, \ldots, K\},\; {t_m} * {s^{(k)}} = x_m^{(k)}.
\end{equation*}

The training consists of learning $(t_m)_{1 \leq m \leq M}$ and $(x_m^{(k)})_{1 \leq k \leq K, 1 \leq m \leq M}$  from the dataset $(s^{(k)})_{1 \leq k \leq K}$. This is performed by solving the optimization problem defined as:
\begin{multline}
\label{eq2}
\minimize{{(t_m)},(x_m^{(k)})} \frac{1}{2}\sum\limits_{k = 1}^K \sum\limits_{m = 1}^M \left( \left\|  {t_m} * {s^{(k)}} - x_m^{(k)}  \right\|_2^2 + \psi\left({x_m^{(k)}} \right) \right)  \\
 \qquad  + \mu \left\| T \right\|_F^2 - \lambda \log \det \left( T \right),
\end{multline}
where $T = \left[ {{t_1}|\ldots|{t_M}} \right]$ concatenates the filters, $\det(T)$ is its determinant\footnote{Throughout the paper, we do not assume that the matrix $T$ is squared. The determinant is computed from its singular value decomposition, following \cite[Prop. 24.68]{Bauschke}}, and $\psi$ penalizes the feature vectors. Equivalently, using matricial notations, one should solve:
\begin{align}
\label{eq3}
\minimize{T,X} \phantom{f} &  \frac{1}{2}\sum\limits_{k = 1}^K \left(\left\| S_k T - {X_k}  \right\|_F^2 + \Psi(X_k) \right)  \nonumber\\
&  + \mu \left\| T \right\|_F^2 - \lambda \log \det \left( T \right).
\end{align}
We introduced the notations: 
\begin{equation*}
 X_k   = \left[ {x_1^{(k)}|\ldots|x_M^{(k)}} \right], \;
S_k T  = \left[ {{t_1} * {s^{(k)}}|\ldots|{t_M} * {s^{(k)}}} \right],
\end{equation*}
with $S_k$ being the Toeplitz matrix associated to the 1D convolution product with filter $s^{(k)}$ (assuming circulant padding). Moreover, $\Psi$ applies the penalty term $\psi$ column-wise on $X_k$, so that $\Psi(X_k) = \sum_{m=1}^M \psi(x_m^{(k)})$, and $X=[X_1^\top|\dots|X_K^\top]^\top$. The CTL strategy uses specific penalization terms over $T$ and $X$. The penalty term on matrix $T$ promotes the diversity among the learnt filters, something which is not guaranteed in CNN. Function $\Psi$ can serve to impose some sparsity prior on $X$ so as to avoid over-fitting \cite{ref17, Briceno2019_coap}. A local minimizer to Problem \eqref{eq3} can be obtained using the proximal alternating algorithm ~\cite{ref19,ref20,ref21}, which alternates between proximal updates on  $T$ and $X$. For more details on the updates derivations and the convergence guarantees, the readers can refer to~\cite{ref1}.

\subsection{Proposed ConFuse Method}
We are now ready to present our novel approach \emph{ConFuse} for the unsupervised construction of representation features of multi-channel data.
A natural strategy is to learn, for each channel $c \in \{1,\ldots,C\}$, a distinct set of convolutional filters $(T^{(c)})_{1 \leq c \leq C}$ and associated features $(X^{(c)})_{1 \leq c \leq C}$, by solving a CTL-based formulation:
\begin{align}
\minimize{T^{(c)},X^{(c)}}  &\frac{1}{2}\sum^K_{k=1} \left(\|S_k^{(c)}T^{(c)}\,-\,X_k^{(c)}\|^2_F + \Psi(X_k^{(c)}) \right) \nonumber \\
& \qquad  +\mu\|T^{(c)}\|^2_F-\lambda\log\det(T^{(c)}).\label{eq:conv}
\end{align}
Then, the learnt channel-wise features are stacked as $X_k=\big[{X_k^{(1)}}^\top | \dots| {X_k^{(C)}}^\top\big]^\top$ for each $k$, and fused by a transform learning procedure acting as a fully-connected layer:%~\cite{ref17}:
\begin{align}
\minimize{\widetilde{T},Z} &\frac{1}{2}\sum\limits_{k=1}^{K}\big\|\widetilde{T}X_k - Z_k\big\|^2_F + \iota_+(Z) \nonumber \\ 
&\qquad + \mu\|\widetilde{T}\|^2_F - \lambda\log\det(\widetilde{T}),\label{eq:fc}
\end{align}
where $\widetilde{T}$ denotes the fusion stage transform (not assumed to be convolutional), $Z$ is the row-wise concatenation of the fusion stage features $(Z_k)_{1\leq k \leq K}$, and $\iota_+$ is the indicator function for positive orthant, equals to zero if all the entries of $Z$ are non-negative, and $+ \infty$ otherwise. Such non-negativity constraint allows to avoid trivial solutions. 

However, the disjoint resolution of Problems \eqref{eq:conv} and \eqref{eq:fc} may lead to unstable solutions that are too sensitive to initialization. We, thus, propose an alternative strategy where we learn all the variables in an end-to-end fashion by solving a joint optimization problem. To this aim, we rely on the key property that the solution $(\widehat{X}^{(c)})_{1 \leq c \leq C}$ of the CTL problem assuming fixed filters $(T^{(c)})_{1 \leq c \leq C}$ can be reformulated as the simple application of an element-wise activation function, that is, for every $k \in \{1,\ldots,K\}$, 
\begin{align}
\widehat{X}_k(T) & = \left[\widehat{X}_k^{(c)}(T)\right]_{1 \leq c \leq C} \nonumber \\
& =  \left[\Phi\big(S_k^{(c)}T^{(c)}\big)\right]_{1 \leq c \leq C},
\label{eq:relu}
\end{align}
with $\Phi$ the proximity operator of $\Psi$ \cite{Combettes}. For example, if $\Psi$ is the indicator function of the positive orthant, then $\Phi$ identifies with the famous rectified linear unit (ReLU) activation function. Many other examples are provided in \cite{Combettes}. Consequently, we propose to plug Equation \eqref{eq:relu} into Problem \eqref{eq:fc}, leading to our final \emph{ConFuse} formulation:
\begin{align}
& \minimize{T,\widetilde{T},Z} \frac{1}{2}\sum\limits_{k=1}^{K}\big\|\widetilde{T} \widehat{X}_k(T) - Z_k\big\|^2_F + \iota_+(Z) + \mu \|\widetilde{T}\|^2_F \nonumber \\ 
& + \mu \|T\|^2_F - \lambda\Big(\log\det(\widetilde{T}) + \sum_{c=1}^C\log\det(T^{(c)})\Big).
\label{eq:joint}
\end{align}
Although Problem \eqref{eq:joint} is still nonconvex, there are two notable advantages with this new formulation. First, we remark that, as soon as the involved activation function is smooth, all terms of the cost function in \eqref{eq:joint} are differentiable, except the indicator function. We can thus employ the accelerated stochastic projected gradient descent, Adam, from \cite{kingma:adam}. %\cite{j.2018on}. 
The latter makes efficient use of automatic differentiation, stochastic approximations to efficiently deal with large-size datasets. Second, any (sub-)differentiable activation function $\Phi$ can be plugged into our model \eqref{eq:relu}, for instance SELU \cite{NIPS2017_6698} or Leaky ReLU \cite{Maas}. This flexibility will play a key role in the performance, as shown in the experimental section. 

\begin{figure}[t]
\includegraphics[width = 3.4in, height = 3.4in]{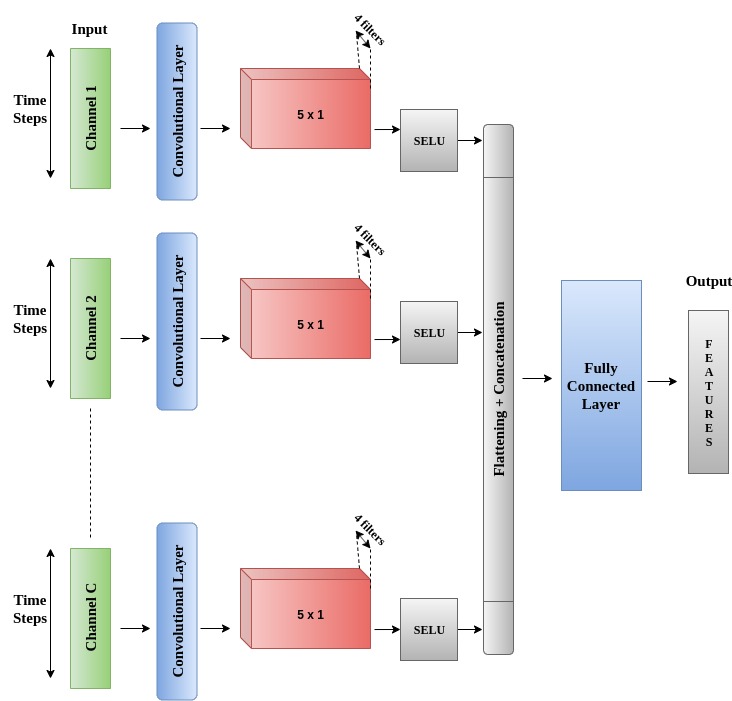}
\caption{Example of \emph{ConFuse} architecture, for $M = 4$ filters of size $5 \times 1$, and SELU activation function.}
\label{fctl}
\end{figure}

An example of structure of the learnt \emph{ConFuse} architecture\footnote{Code available at -  https://github.com/pooja290992/ConFuse} is shown in Figure \ref{fctl}. Note that our approach is completely unsupervised. Specifically, we replaced supervision by explicitly learning the features $Z$, on which we impose the non-negativity constraint to avoid trivial solutions. Regarding the representation filters stacked in matrices $(T,\widetilde{T})$, %the logarithm determinant regularization forces diversity among them, whereas the Frobenius regularization ensures that their coefficients stay bounded. 
the log-det regularization imposes a full rank on those. Thus, it helps to enforce the diversity and to prevent the degenerate solution ($ T = 0, X = 0, \widetilde{T} = 0, Z = 0 $). The Frobenius regularization ensures that the matrices entries remain bounded.

\section{EXPERIMENTAL EVALUATION}
\label{sec:experiment}
We evaluate the proposed method, \emph{ConFuse}, on the real-world problems of stock price prediction, and trading under stock forecasting. Both can be formulated as a multi-channel time series analysis, where the inputs are $C = 5$ raw variables: opening price, closing price, (day’s) low(est) price, (day’s) high(est) price, and net asset value (NAV). As financial data are 1D signals, we will make use of 1D convolutional filters in our CTL representation, and the learnt features will then be fused to produce the final representation. %The architecture of the ConFuse network is summarized in Table  \ref{tab1}.

Experiments are carried out on the data from the National Stock Exchange (NSE) of India. The dataset contains information of 150 stocks between 2014 and 2018. Companies available are from various sectors, such as information technology (TCS, INFY), automobile (HEROMOTOCO, TATAMOTORS), banking (HDFCBANK, ICICIBANK).

We have compared \textit{ConFuse} with two state-of-the-art deep learning based time series models, namely TimeNet \cite{ref15} based on LSTM, and ConvTimeNet \cite{ref16} based on CNN. These models were originally developed for univariate data. They were converted to multivariate form owing to our requirement of feeding five types of inputs. Each of the inputs is processed by either TimeNet or ConvTimeNet and the features from the five channels are fused by a fully connected layer with an output target. For the price forecasting problem, the target is multivariate – the outputs are the next day’s variables (open, close, high, low and NAV). Since the outputs are real, the activation function at the output is linear. For the stock trading problem, the output is a binary BUY/SELL. Softmax is used to get the labels after the fully connected layer. The architectures of \textit{ConFuse}, ConvTimeNet, and TimeNet are summarized in Table \ref{tab1}. %\textcolor{red}{
We have tested six different activation functions for \emph{ConFuse}.\footnote{The gradients of SELU, RELU, PRELU, and Leaky RELU are not defined in zero. It is customary to consider any valid sub-gradient value instead. We did not notice any practical convergence issues with ADAM algorithm when resorting to this strategy.}

The optimizer used for each architecture is Adam %\cite{j.2018on} 
with parameters $(\beta_1,\beta_2) = (0.9,0.999)$ and $\epsilon = 10^{-8}$. Weight decay is set to $5 \times 10^{-5}$ for \emph{ConFuse} and TimeNet, and $10^{-4}$ for ConvTimeNet. The regularization parameters $\mu = 10^{-4}$ and $\lambda = 10^{-2}$ are used in \emph{ConFuse}, for all activation functions. These configurations give the best results. 

The proposed technique, \emph{ConFuse}, generates unsupervised features. In the forecasting problem, we input these features to ridge regression for forecasting. Evaluation is carried out in terms of mean absolute error (MAE) between the predicted and actual stock prices. The averages of MAE of all the stocks forecasting results are shown in Table \ref{tab2}. For the stock trading problem, we apply a random decision forest to the generated features for classification. We evaluate the results in terms of area under the ROC curve (AUC), precision, recall, F1 score and relative difference (between actual and predicted) in annualized return (AR). 

\begin{threeparttable}
\caption{\textbf{Description of Different Techniques}}\label{tab1}
\small
\begin{tabular}{|l|p{5.9cm}|}
\hline
\textbf{Method} &  \textbf{Architecture Description}\\
\hline
ConFuse & 
$5 \times \begin{cases}
\textbf{layer1}\colon\, \textbf{1D Conv} (1,4,5,1,2)^\dagger\\
         \textbf{Activation}\;\text{(e.g., ReLU)}\\
\end{cases}$
\newline
\newline
1 $\times$ \textbf{layer2}: \textbf{Transform Learning}%\\%\newline
\\
\hline
ConvTimeNet & 
\( 
5 \times \begin{cases}
\textbf{layer1}: \textbf{1D Conv}
         (1,32,9,1,4)^\dagger\\
         \textbf{Batch Norm} + 
         \textbf{SELU}\\
\textbf{layer2}: \textbf{1D Conv}
         (32,32,3,1,1)^\dagger\\
         \textbf{Batch Norm} + 
         \textbf{SELU} + \textbf{Skip conn.}\\
\textbf{layer3}: \textbf{1D Conv}
         (32,64,9,1,4)^\dagger\\
         \textbf{Batch Norm} + 
         \textbf{SELU}\\
\textbf{layer4}: \textbf{1D Conv}
         (64,64,3,1,1)^\dagger\\
         \textbf{Batch Norm} + 
         \textbf{SELU} +\textbf{Skip conn.}\\
\textbf{layer3}: \textbf{Global Average Pooling}\\
\end{cases}\)
\newline
\newline
\textbf{layer4}: \textbf{Fully Connected}
\newline 
\textbf{For Trading, added} \textbf{layer5} : \textbf{Softmax}
\\
\hline
TimeNet & 
$5 \times \begin{cases}
\textbf{layer1} : \textbf{LSTM unit}\\
(1,12,2,\text{True})^\mathsection\\
\textbf{layer2} : \textbf{Global Average Pooling}\\
\end{cases}$
\newline
\newline
\textbf{layer3} : \textbf{Fully Connected}
\newline 
\textbf{layer4} : \textbf{Softmax}
\\
\hline
\end{tabular}
\begin{tablenotes}
\item[$\dagger$] \small{input bands, output bands, kernel size, stride, padding}
\item[$\mathsection$] \small{input size, hidden size, num. layers, bidirectional}
\end{tablenotes}
\end{threeparttable}
% \end{table}
% \begin{table}%[!htb]
%\begin{threeparttable}
\begin{table}[!htb]
\caption{\textbf{Forecasting Results (MAE)}}\label{tab2}
\small\centering
\begin{tabular}{|p{2.85cm}@{\,}|c@{\quad}c@{\quad}c@{\quad}c@{\quad}c|}
\hline
\textbf{Method} & \textbf{Open} & \textbf{Close} & \textbf{High} & \textbf{Low} & \textbf{NAV}\\
\hline
ConFuse-SELU & 0.011 & 0.023 & 0.017 & 0.017 & 0.447 \\
\hline
ConFuse-ReLU & 0.009 & 0.021 & 0.014 & 0.014 & 0.445\\
\hline
ConFuse-PReLU & 0.007 & 0.017 & 0.012 & 0.013	& 0.434\\
\hline
ConFuse-LeakyReLU & \textbf{0.007} & \textbf{0.017} & \textbf{0.012} & \textbf{0.013} & \textbf{0.427}\\
\hline
ConFuse-Tanh & 0.258 & 0.259 & 0.258 & 0.259 & 0.488\\
\hline
ConFuse-Sigmoid & 0.227 & 0.227 & 0.227 & 0.227 & 0.482\\
\hline
ConvTimeNet & 1.551 & 1.554 & 1.535 & 1.567	& 2.357\\
\hline
TimeNet	& 0.295 & 0.295 & 0.294 & 0.296 & 0.511 \\

% ConFuse & \textbf{0.01} & \textbf{0.02} & \textbf{0.01} &
% \textbf{0.01} & \textbf{0.44} \\
% \hline
% ConvTimeNet & 1.55 & 1.55 & 1.53 & 1.56 & 2.35 \\
% \hline
% TimeNet & 0.30 & 0.30 & 0.29 & 0.30 & 0.51 \\
\hline
\end{tabular}
\end{table}
%\end{threeparttable}
% \end{table}
% \begin{table}%[!htb]
%\begin{threeparttable}
\begin{table}[!htb]
\caption{\textbf{Trading Results}}\label{tab3}
\small\centering
\begin{tabular}{|p{2.85cm}@{\,}|c@{\;\;}c@{\;\;}c@{\quad}c@{\,}c@{\,}|}
\hline
\textbf{Method} & \textbf{Precis.} & \textbf{Recall} & \textbf{F1} & \textbf{AUC} & 
\textbf{AR}  \\
\hline
ConFuse-SELU & \textbf{0.524}	& \textbf{0.777} & \textbf{0.619} & \textbf{0.543} & \textbf{17.898}
\\
\hline
ConFuse-RELU & 0.505 & 0.648 & 0.556	& 0.523 & 18.112
\\
\hline 
ConFuse-PRELU & 0.491 & 0.601 & 0.528 & 0.506 & 19.091 \\
\hline 
ConFuse-LeakyRELU & 0.496 & 0.602 & 0.531 & 0.511 & 19.150 \\
\hline
ConFuse-Tanh & 0.469 & 0.560 & 0.493 & 0.497 & 19.002\\
\hline 
ConFuse-Sigmoid & 0.487 & 0.584 & 0.513 & 0.498 & 20.540\\
\hline
ConvTimeNet & 0.457 & 0.507 & 0.413 & 0.524 & 19.410
\\
\hline
TimeNet	& 0.469 & 0.648 & 0.496 & 0.513 & \begin{tabular}[t]{c}18.764\end{tabular}\\

\hline
\end{tabular}
%\end{threeparttable}
\end{table}

\begin{figure}
%\includegraphics[width = 3.5in, height = 1.7in]{images/jswenergy.png}
%\caption{Stock Forecasting Performance}%\label{close_plots}
% \end{figure}
% \begin{figure}%[!htb]
\includegraphics[width = 3.5
in, height = 1.7in]{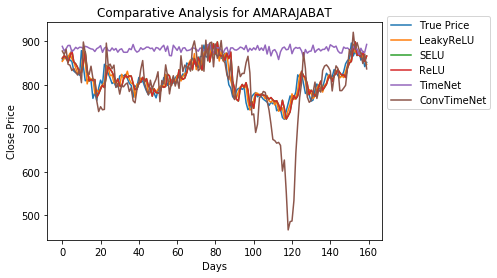}
\caption{Stock Forecasting Performance}
\label{close_plots}
\end{figure}

\begin{figure}
\includegraphics[width = 3.5in, height = 1.4in]{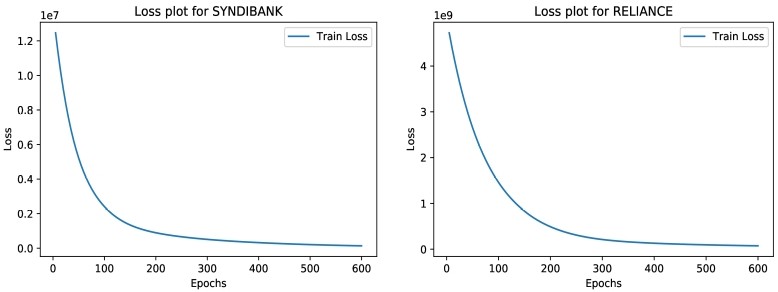}
\caption{Empirical Convergence Plots}\label{loss_plots}
\end{figure}

We find that the results for the stock forecasting problem are exceptionally good. For most tested activation functions, \emph{ConFuse} has MAE more than one order of magnitude lower than the state-of-the-arts. %\textcolor{red}{for every tested activation function}. 
For the stock trading problem, \emph{ConFuse} outperforms the benchmarks as well %\textcolor{red}{
with two activation functions, namely SELU and ReLU, and reaches similar performance than the benchmarks with the other activation functions. Given the limitations in space, we can only show the average values over the 150 stocks. For the qualitative evaluation of a few stocks, we display example of forecasting result on a randomly chosen stock in Figure \ref{close_plots}, which corroborates the very good performance of the proposed solution. We also show some empirical convergence plots of Adam in Figure \ref{loss_plots} when using \emph{ConFuse} with SELU, which depict the practical stability of our end-to-end training method.

\section{CONCLUSION}
\label{sec:conclusion}
This work proposes \emph{ConFuse}, a novel multi-channel unsupervised fusion framework based on learnt convolutional filters. It overcomes the fundamental issue of convolutional nets, that is their inability to learn in an unsupervised fashion. Our work is based on the premise of CTL \cite{ref1}, a technique for learning convolutional filters in an unsupervised fashion. In the proposed fusion framework, each of the channels is processed by CTL. The outputs from these filters are concatenated and fused by fully-connected transform learning. An original end-to-end training strategy is employed. We find that the proposed technique improves over state-of-the-art deep learning based time series analysis techniques. Currently, our model only uses one layer of learnt convolutional filters. In future work, we would like to extend this work to a multi-layer architecture.

\bibliographystyle{IEEEtran}
\bibliography{references.bib}

\end{document}